\documentclass[letter,10pt,conference]{IEEEtran}
\IEEEoverridecommandlockouts

\usepackage{amsmath,amssymb,amsfonts}
\usepackage{graphicx}
\usepackage{textcomp}
\usepackage{xcolor}
\usepackage{tabularx}
\usepackage{color, colortbl}

\definecolor{lime}{rgb}{0.88,2,10}
\usepackage{subcaption}
\usepackage{wrapfig}
\usepackage[linesnumbered,ruled,vlined]{algorithm2e}
\usepackage{xpatch}
\usepackage{url}
\usepackage{multirow}
\usepackage{multicol}
\usepackage{wrapfig}
\usepackage[export]{adjustbox}
\usepackage[numbers,sort&compress]{natbib}
\usepackage{nomencl}
\usepackage{siunitx}
\usepackage{blindtext}
\usepackage[T1]{fontenc}
\usepackage[spaces,hyphens]{xurl}
\usepackage{float}

\usepackage{fancyhdr}
\fancypagestyle{mystyle}{
    \chead{2022 IEEE International Conference on Big Data (IEEE BigData 2022)}
}

\SetAlFnt{\small}

\SetKwComment{Comment}{$\triangleright$\ }{}

\usepackage[utf8]{inputenc}
\usepackage{enumitem}
\SetKwInput{kwInit}{Initialize}

\definecolor{lime}{rgb}{0.5, 1.0, 0.8}

\def\BibTeX{{\rm B\kern-.05em{\sc i\kern-.025em b}\kern-.08em
    T\kern-.1667em\lower.7ex\hbox{E}\kern-.125emX}}

\usepackage[numbers,sort&compress]{natbib}

\newcommand{\fref}[1]{Fig.~\ref{#1}}

\newcommand{\sref}[1]{Section~\ref{#1}}

\newcommand\HUGE{\fontsize{22.5}{30}\selectfont}
\begin{document}
\title{\HUGE AsyncFLEO: Asynchronous Federated Learning for LEO Satellite Constellations with High-Altitude Platforms}

\author{Mohamed Elmahallawy and Tie Luo$^*$\thanks{* Corresponding Author.}\\
{Computer Science department, Missouri University of Science and Technology, USA}\\
{E-mail: \{meqxk, tluo\}}@mst.edu}

\maketitle
\thispagestyle{mystyle}

\begin{abstract}
Low Earth Orbit (LEO) constellations, each comprising a large number of satellites, have become a new source of big data ``from the sky''. Downloading such data to a ground station (GS) for big data analytics demands very high bandwidth and involves large propagation delays. Federated Learning (FL) offers a promising solution because it allows data to stay in-situ (never leaving satellites) and it only needs to transmit machine learning model parameters (trained on the satellites' data). However, the conventional, synchronous FL process can take several days to train a single FL model in the context of satellite communication (Satcom), due to a bottleneck caused by {\em straggler satellites}. In this paper, we propose an asynchronous FL framework for LEO constellations called {\em AsyncFLEO} to improve FL efficiency in Satcom. Not only does AsynFLEO address the bottleneck (idle waiting) in synchronous FL, but it also solves the issue of {\em model staleness} caused by straggler satellites. AsyncFLEO utilizes high altitude platforms (HAPs) positioned ``in the sky'' as parameter servers, and consists of three technical components: (1) a {\em ring-of-stars} communication topology, (2) a {\em model propagation} algorithm, and (3) a {\em model aggregation} algorithm with {\em satellite grouping} and {\em staleness discounting}. Our extensive evaluation with both IID and non-IID data shows that AsyncFLEO outperforms the state of the art by a large margin, cutting down convergence delay by 22 times and increasing accuracy by 40\%. 
\end{abstract}


\section{Introduction}\label{sec:intro}


Recent years have seen a surge in the deployment of Low Earth Orbit (LEO) satellites, which float at an altitude of 500--2000 km and form multiple (mega) constellations. Many of these satellites continuously collect Earth observational data in high volume, speed, and heterogeneity, constituting a new source of big data ``from the sky''. To extract tremendous value from such data and thereby support a wide range of applications such as urban planning, weather forecasting, and disaster management \cite{wu2020resource, perez2021airborne}, machine learning (ML) plays a crucial role in big data analytics. However, downloading a massive amount of data (e.g., millions of satellite images) to a ground station (GS) to train ML models is not practical: 1) satellite communication (Satcom) can barely afford such a high demand on network bandwidth; 2) the propagation delay between LEO satellites and GS is large; 3) transmission of raw data involves significant privacy and security risks (e.g., for military applications). 

Federated learning (FL) \cite{fl2021} offers a promising solution by allowing each satellite to train an ML model {\em locally} on its own data and only need to upload the resulting model parameters to a parameter server (PS). The PS then aggregates all the satellites' {\em local models} into a {\em global model}. This eliminates the need for transmitting raw data, thus coping with the bandwidth and privacy issues mentioned above quite well. 

However, applying FL to Satcom or more specifically LEO constellations is not straightforward and faces significant challenges. First, FL is an iterative process and typically requires hundreds of communication rounds between clients (i.e., satellites) and the PS. While this does not present as a big issue in large-capacity networks, it causes a severe slow-convergence problem in the context of Satcom due to the long propagation and transmission delay, and more saliently because of the {\em highly sporadic and irregular visit pattern} of LEO satellites to the PS. This visit pattern results from the distinction between satellites' and the PS' travel trajectories 
and the distinction between satellite orbiting speeds and the Earth rotation speed. Ultimately, the iterative FL process will incur vast delay in Satcom, taking several days or even longer to converge \cite{so2022fedspace,razmi}.




To address this severely slow convergence problem of FL when applying to Satcom, we propose {\em AsyncFLEO}, a novel asynchronous FL framework tailored for LEO satellites, in this paper. AsyncFLEO exploits the availability of satellite local models opportunistically without waiting for all the models to be available (as in synchronous FL), and tackle {\em straggler satellites} and {\em model staleness} as follows. In the first epoch, it groups satellites from various orbits based on their data distributions (which we infer using their model weights since data is not accessible to PS in FL). Then in subsequent epochs, the PS selects a subset of satellite models from each group based on their freshness, to be included in model aggregation. Apart from these, AsyncFLEO has two other technical components for efficiency improvement: a ring-of-stars communication topology and an intra-orbit model propagation algorithm. Overall, AsyncFLEO can accelerate the FL convergence speed by several orders of magnitude, while increasing model quality (in terms of classification accuracy) at the same time.

AsyncFLEO leverages high altitude platforms (HAPs) positioned ``in the sky'' as PSs in lieu of a GS \cite{happaper}. A HAP is a semi-static aircraft or airship situated in the stratosphere (17--22 km above the Earth's surface) and thus offers slightly better visibility of satellites than locating the PS on the ground \cite{happaper,hsieh2020uav,arum2020review}. 
However, AsyncFLEO does not {\em rely on} HAPs and it works with GS exactly the same way as in its single-HAP case.

In summary, this paper makes the following  contributions:
\begin{itemize}[leftmargin=*]
    \item We propose for LEO Satcom an asynchronous FL approach that overcomes the large convergence delay caused by the highly sporadic connectivity and irregular visit pattern between satellites and the FL server.

    \item AsyncFLEO introduces three new technical components: (1) a model aggregation algorithm based on satellite grouping and model selection, which tackles {\em straggler satellites} and {\em stale models}; (2) a {\em ring-of-stars} topology in substitution of the conventional FL's star topology; (3) a {\em model propagation algorithm} that alleviates sporadic connectivity using intra-orbit model relay.

    \item  We demonstrate via simulations that AsyncFLEO significantly accelerates FL convergence and outperforms state-of-the-art FL-Satcom algorithms by large margins (up to 22 times faster in convergence and 40\% increase in accuracy).
\end{itemize}



\noindent\textbf{Paper Organization.} Section~\ref{Sec:related_work} discusses related work.  Section~\ref{sec:model} describes a general system model for FL in LEO constellations with HAPs. Section~\ref{sec:AsyncFLEO} describes the design of AsyncFLEO in great detail. 
The performance evaluation of AsyncFLEO with other state-of-the-art methods is provided in Section \ref{sec:simu}. Finally, we conclude in Section \ref{section 4}.

\section{Related Work}\label{Sec:related_work}
While the research on big data problems in FL-Satcom is still in its infancy, some initial progress has been made recently by a number of interesting studies \cite{chen2022satellite,razmi2022ground, happaper, so2022fedspace, razmi}.

\textbf{Synchronous FL.} Chen et al. \cite{chen2022satellite} applied the traditional FL approach (i.e., FedAvg \cite{mcmahan2017communication}) to LEO constellations and compared it with centralized training obtained from downloading the data to a GS. 
Razmi et al. \cite{razmi} proposed an FL approach called FedISL which uses inter-satellite link (ISL) for communication among satellites within the same orbit to reduce FL delay. It assumes that the PS is either a GS located at the North Pole (NP) or a satellite in medium Earth orbit (MEO) at an altitude of 20,000 km, in order to simplify the problem and increase the visibility between satellites and PS. However, NP is the ideal location which was only available during the early years of LEO deployment, and PS at MEO of that altitude travels at very high speed with non-negligible {\em Doppler shift} which is not addressed in their study. 
FedHAP \cite{happaper} is an alternative synchronous FL approach proposed to address the above limitations. It introduces HAPs in place of the traditional GS, to act as PSs to oversee the model aggregation process collaboratively. However, FedHAP \cite{happaper} still requires more than a day to converge due to its synchronicity.

\textbf{Asynchronous FL.} The authors of \cite{razmi2022ground} proposed an asynchronous FL approach called FedSat to speed up FL convergence. Similar to \cite{razmi}, it also assumes the ideal setup where the GS is located at the NP, so that every satellite visits the GS at regular intervals, which again over-simplifies the problem. 
So et al. \cite{so2022fedspace} proposed an approach called FedSpace to trade-off between the idle connectivity in synchronous FL and the staleness issue in asynchronous FL. 
The limitation of FedSpace is that it requires each satellite to upload a fraction of its captured images (or all the images in a lower resolution) to the GS to schedule model aggregation, which contradicts the important FL principle on privacy protection and communication efficiency.


\section{System Model} \label{sec:model}

\begin{figure}[ht]
     \centering
     \includegraphics[width=1\linewidth]{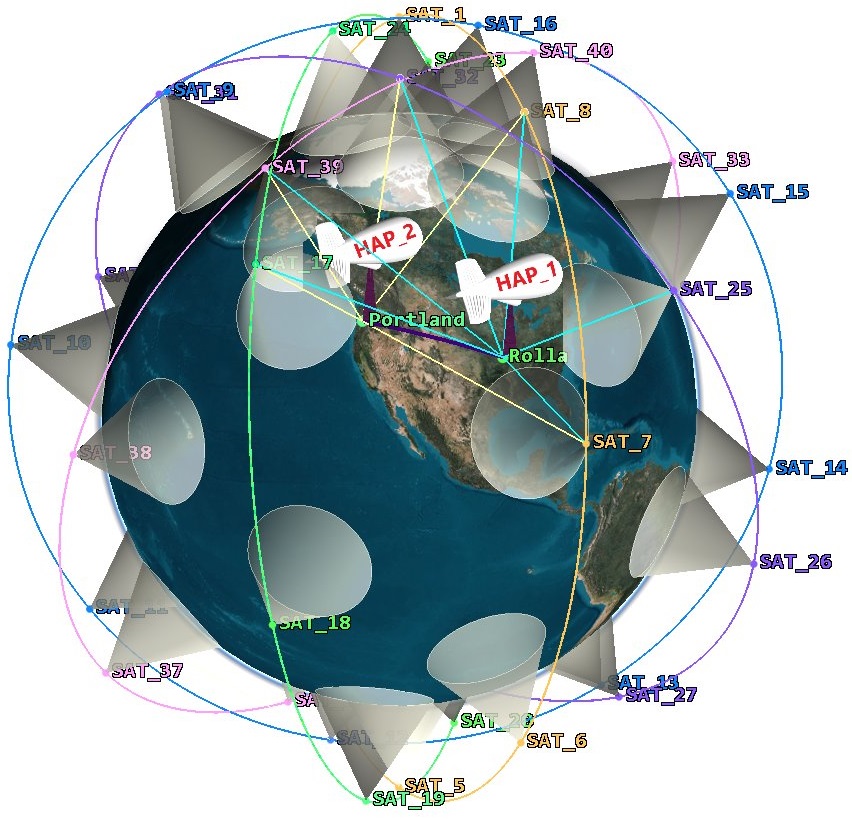}
    \caption{An example of Walker-delta constellation \cite{walker1984satellite} consists of $O = 5$ orbits, each having $N_{o}= 8$ satellites, orchestrated by $H = 2$ HAPs. Gray cones depict LEO satellites' coverage.}
    \label{Picture1}
\end{figure}

Consider an LEO satellite constellation with $\textit{O}$ orbits where each orbit $\textit{o}$ contains $N_{o}$ satellites equally spaced. 
Any satellite in orbit $o$ has an orbital period $T_{o}= \frac{2 \pi (R_{E}+h_{o})}{v_{o}}$, where $R_{E}=6371$km is the radius of the Earth, $h_{o}$ is the orbital altitude, and $v_{o}$ is the satellite velocity given by $v_{o}=\sqrt{\frac{GM}{(R_{E}+h_{o})}}$, in which $G$ is the gravitational constant and $M$ is the mass of the Earth. 
There are a few HAPs ${\mathcal H}=\{h_1, h_2,\dots, H\}$ that act as PSs and communicate with a varying number of satellites at different times depending on the {\em irregular} visit pattern. As mentioned in \sref{sec:intro}, AsyncFLEO does not rely on HAPs and can operate with GS the same way as the single-HAP case; we include HAPs just to present a better design scheme (slightly better visibility than GS due to its elevated altitude, as well as more flexibility since it is deployed at a fixed location ``in the sky'').
In fact, our performance evaluation in \sref{sec:simu} covers both HAP and GS scenarios with AsyncFLEO. 
Fig. \ref{Picture1} gives an illustration.

\subsection{FL in Satellite Communication Networks}\label{sec:flsat-basics}
\begin{figure*}[t]
     \centering
     \includegraphics[width=\linewidth]{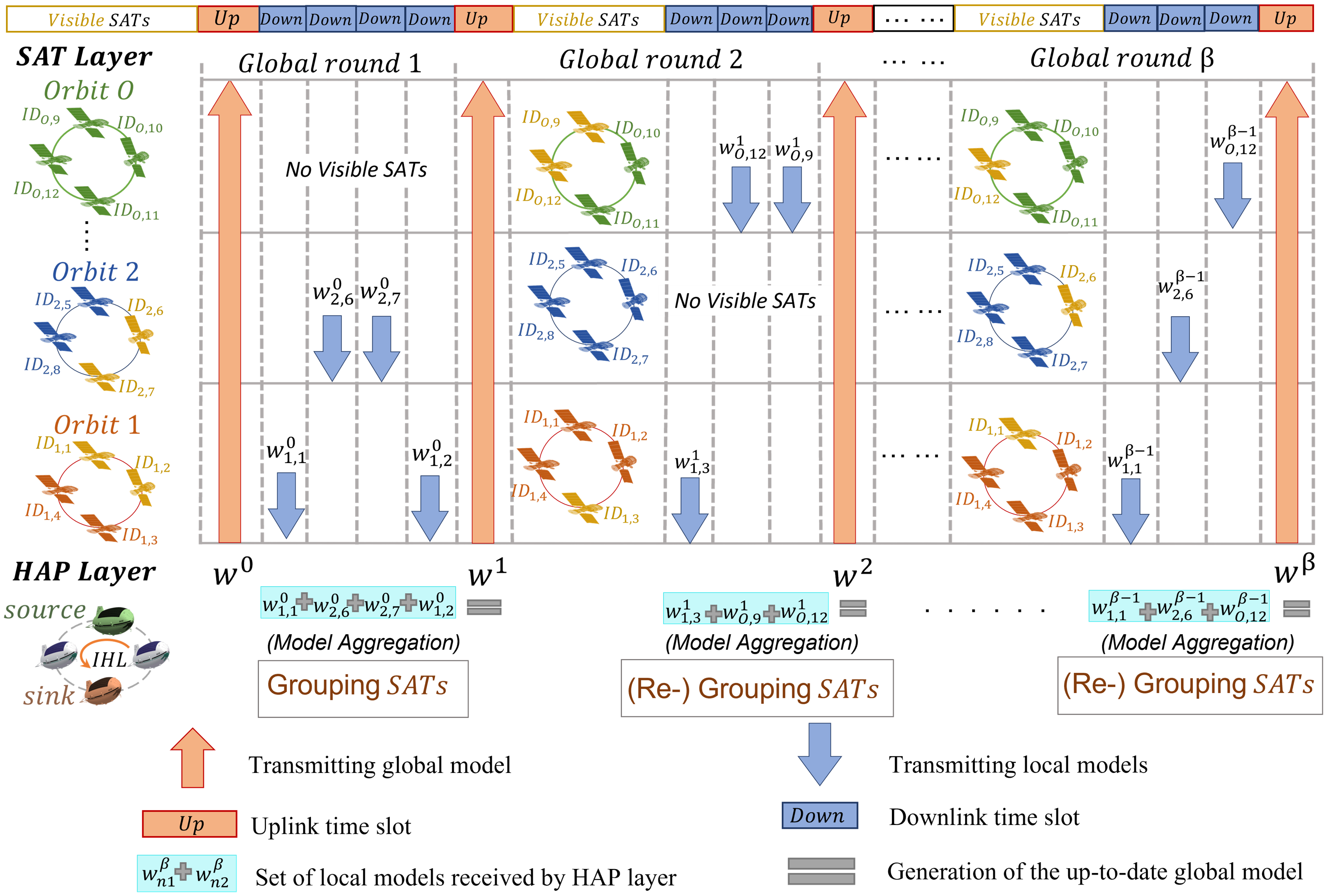}
    \caption{The AsyncFLEO framework illustrated using a sequential diagram. {\bf Yellow} satellites represent {\bf visible} satellites.}
    \label{Picture5}
\end{figure*}
In this section, we assume a single PS (either HAP or GS) for ease of description and more consistency with standard FL. With this background knowledge, it would lead to a more natural flow to \sref{sec:AsyncFLEO}, where we present a more sophisticated scenario with multiple HAPs.

The overall goal of FL for a constellation $\mathcal{N}$ of LEO satellites is to collaboratively train a global ML model under the orchestration of a PS, using each satellite's data {\em locally}, by minimizing the following global objective function:
\begin{equation} \label{eqn1}
    \begin{array}{rrclcl}
           \displaystyle \min_{w \in \mathbb{R}^{d}} F(w)= \sum_{n\in \mathcal{N} }{\frac{m_{n}}{m}{F_{n}{(w)}}}
    \end{array}
\end{equation}
where $w$ is the parameters of the target global model, $m_{n}$ is the size of satellite $n$'s dataset $D_{n}$, ${m} = \sum_{n\in \mathcal{N}} m_{n}$ is the total size of all the satellites' data, and $F_{n}$ is the local loss function at satellite $n$ resulting from training $w$ over $D_{n}$, which can be expressed as
\begin{equation}\label{eqn2}
     F_{n}{(w)} = \frac{1}{m_{n}}\sum_{x\in D_{n}} f_{n}(w;x) 
\end{equation}
where $f_{n}(w;x)$ is the training loss for a data sample \textit{x} and model parameters $w$ at satellite $n$. Each satellite $n$ solves \eqref{eqn2} by applying a local optimizer such as stochastic gradient descent (SGD), for $J$ local iterations, in the following way:
\begin{equation}
    w_{n}^{	\beta,j+1} = w_{n}^{\beta,j}- \frac{\eta}{b} 
    \sum_{i=1}^b \nabla f_{n}(w_{n}^{\beta,j};x_{n}^{i}) 
\end{equation}
where $j=0,1,2,...,J-1$ is a local training iteration in a global training round $\beta=0,1,2,\dots$, $w_{n}^{\beta,j}$ is the local model of satellite $n$ at iteration $j$, $\eta$ is the learning rate, $x_{n}^{i}$ is the $i$-th training sample in the current mini-batch of size $b$. 

In general, there are two approaches for the PS to aggregate all the local ML models $w_{n}^{\beta,J}$ collected from the satellites $\mathcal N$, into a global model. {\em Synchronous FL} follows the same principle as McMahan et al. \cite{mcmahan2017communication}, in which the PS waits to receive all the satellites' local models and then aggregates them as: 
\begin{equation}\label{eq:fedavg}
    w^{\beta+1}= \sum_{n\in\mathcal N} \frac{m_{n}}{m} w_{n}^{\beta}
\end{equation}
where we write $w_{n}^{\beta}$ in place of $w_{n}^{\beta,J}$ for notation simplicity. Since this approach requires all the satellites in a constellation to become successively visible to the PS, 
it incurs a large delay in each global communication round and hence a much-prolonged convergence time after all the rounds.

{\em Asynchronous FL} is an approach 
that aims to address this limitation in synchronous FL. With this approach 
\cite{xie2020asynchronous}, the PS aggregates just a subset of local models as soon as they have been received, 
thus mitigating the long idle waiting as in synchronous FL and speeding up the global model convergence. However, it introduces a {\em model staleness} problem, where some received models could come from earlier rounds from {\em straggler satellites} with limited visibility and hence are outdated. According to \cite{so2022fedspace}, this problem makes asynchronous FL unable to achieve comparable accuracy to synchronous FL although the convergence time would be reduced; in other words, one has to make a trade-off. In contrast, our proposed asynchronous FL approach AsyncFLEO is able to accelerate convergence {\em and} improve accuracy simultaneously. AsyncFLEO determines the subset of satellites of each round based on their data distributions (inferred from their model weights) using a satellite grouping scheme, and uses a staleness discounting factor to progressively aggregate models as soon as a model becomes available (details are given in Section \ref{sec:group-agg}).


\subsection{Communication Links}\label{sec:model-links}



Without loss of generality, consider a satellite \textit{n} and a PS \textit{g}, where the communication link between them can only be established if $\vartheta_{n,g}(t) = \angle (r_{g}(t), (r_{n}(t) - r_{g}(t))) \leq \frac{\pi}{2}-\vartheta_{min}$, where $r_{n}(t)$ and $r_{g}(t)$ are the trajectory of satellite \textit{n} and GS \textit{g}, respectively, and $\vartheta_{min}$ is the minimum elevation angle (a constant depending on the device). 

Below, we model all the communication links (among satellites or HAPs or between them) as radio frequency (RF) links rather than free-space optical (FSO) links for the purpose of a fair comparison with prior work, but we note that, in practice, AsyncFLEO can actually benefit from FSO links which enjoy a much higher data rate (as high as Terabytes per second) than RF links and are also much more resistant to radio interference. 

Assuming that the wireless channels are symmetric with additive white Gaussian noise, then the signal-to-noise ratio (SNR) between two objects $x$ and $y$ (e.g., satellite and GS) in free space is given as 
\begin{equation}
    SNR_{RF}(x,y) = \frac{P_{t}G_{x}G_{y}}{K_{B} T B \mathcal{L}_{x,y}}  
\end{equation}
where $P_{t}$ is the transmission power, $G_{x}$ and $G_{y}$ denote the total antenna gain of the transmitter and the receiver, respectively, $K_{B}$ is the Boltzmann constant, \textit{T} is the noise temperature at the receiver, \textit{B} is the channel bandwidth, and $\mathcal{L}_{x,y}$ is the free-space pass loss which can be expressed as
\begin{equation}
    \mathcal{L}_{x,y} =
     \begin{cases}
      \Big(\frac{4\pi \|x,y\|_{2}  f}{c}\Big)^{2}, & \text{if LoS($x$, $y$),} \\ 
      ~~~~~~\infty ~~~~~~~ , & \text{otherwise}.
     \end{cases}
\end{equation}
where $\|x,y\|_{2}$ is the Euclidean distance between $x$ and $y$, $f$ is the carrier frequency, \textit{c} is the speed of the light, and LoS is the line-of-sight link between $x$ and $y$. 
The total delay $t_{c}$ of sending data from $x$ to $y$ or vice versa, can be computed as follows:
\begin{align}
    t_{c} &= t_{t}+t_{p}+t_{x}+ t_{y} \label{eqnx} \\
    t_{t} &= \frac{b{|\mathcal{D}|}}{R} , \quad t_{p} = \frac{\|x,y\|_{2}}{c} \label{eqny}
\end{align}
where $t_t$ is the transmission delay, $t_p$ is the propagation delay, $t_{x}$ and $t_{y}$ are the processing delay at $x$ and $y$, respectively,  ${|\mathcal{D}|}$ is the number of data samples and $b$ is the size in bits of each sample, and $R$ is the data rate that can be given by the Shannon formula as
\begin{equation} \label{eqnz}
    R \approx B \log_{2} (1+SNR)
\end{equation}

In our simulation (\sref{sec:simu}), we set the parameters for the formulae presented above.

\section{AsyncFLEO} \label{sec:AsyncFLEO}

AsyncFLEO is an asynchronous FL approach tailored for LEO satellites to speed up FL model convergence without sacrificing the model performance. It achieves this by (1) alleviating the negative impact of stale models received from straggler satellites due to irregular and sporadic connectivity, (2) overcoming the long waiting time for each satellite to visit the PS, and (3) reducing the large number of communication rounds via a smarter selection of satellites in each round. AsyncFLEO consists of three technical components described in subsections A--C: (A) a new, {\em ring-of-stars} topology for communication between satellites and HAPs, in lieu of the {\em star topology} used in traditional FL; (B) a {\em model propagation} algorithm that relays local and global ML models among satellites (intra-orbit) and HAPs; (C) a {\em model aggregation} algorithm based on satellite grouping and a smarter model selection of the fresh satellite models from each group to be included in each asynchronous round
as well as a {\em discounting} scheme designed for and applied to stale models. Fig.~\ref{Picture5} gives an overview of the AsyncFLEO framework. 

\subsection{SAT-HAP Communication Topology}

\begin{figure}[ht]
     \centering
     \includegraphics[width=1\linewidth]{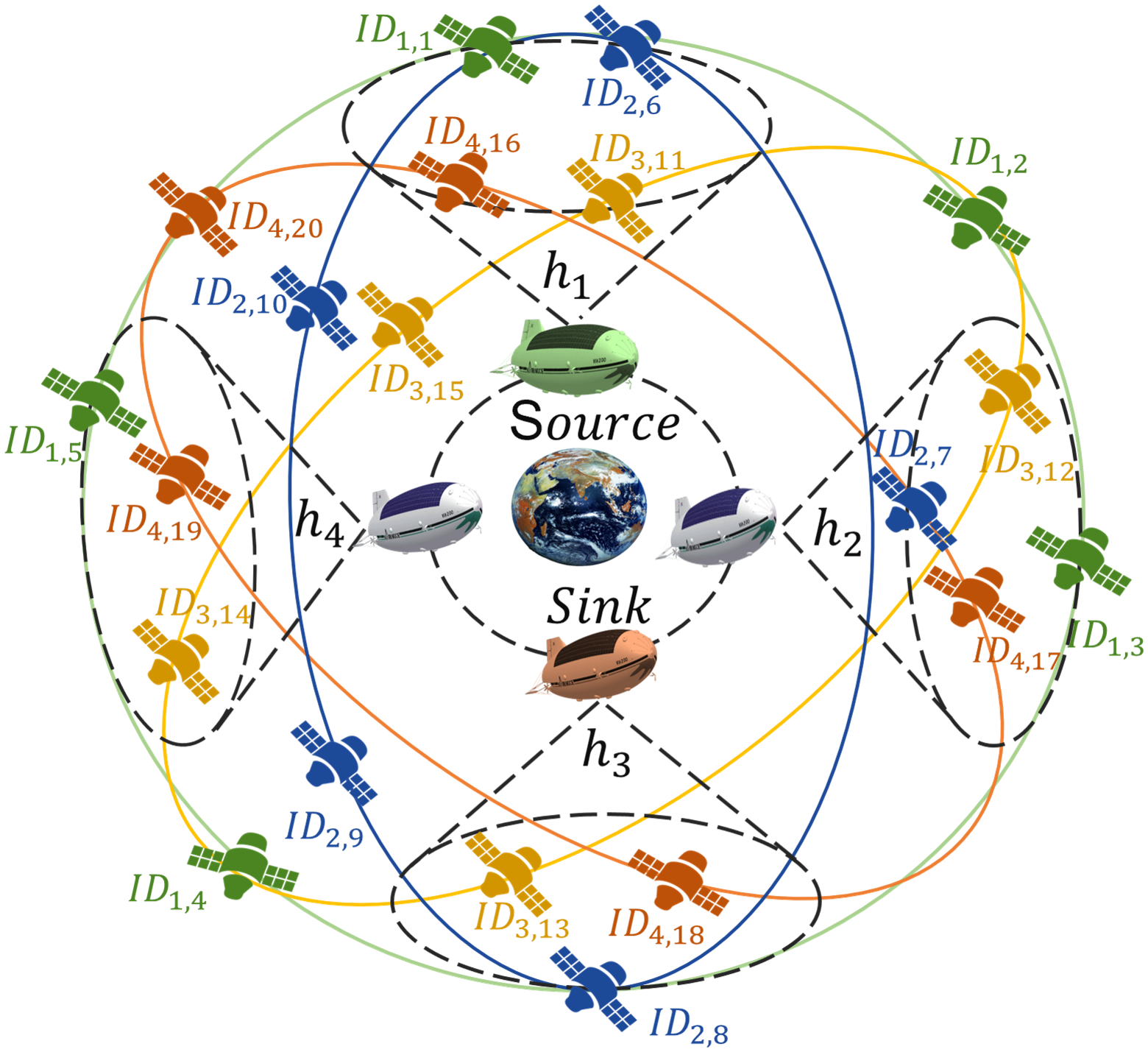}
    \caption{Illustration of the {\em ring-of-stars} topology (indicated by the black dotted lines). The 4 HAPs form the ``backbone'' ring and each HAP orchestrates a star topology consisting of its currently visible satellites (as in the cone). Satellite IDs are in the format of ($ID_{Orbit\#,~ Satellite\#}$). The large colored ovals represent orbits.}
    \label{fig:topo}
\end{figure}

The conventional FL adopts a star topology, where a PS communicates with all the clients (satellites in our case) directly. 
In AsyncFLEO, we introduce parallelism by using a ring-of-stars topology as follows. First, we cast the network into a layered structure where the first layer is an SAT layer that consists of all the LEO satellites $\mathcal N$, and the second layer is a HAP layer that consists of all the HAPs $\mathcal H$ (or a GS, which is just a special case). In the HAP layer, HAPs form a {\em ring topology} in which they can communicate with their adjacent (two) neighbors; at the same time, each HAP can also communicate with all of its currently visible satellites from {\em different orbits}, thereby forming a (small) star topology. Thus overall, all the HAPs and their currently visible satellites form a {\em ring-of-stars} topology, as shown in \fref{fig:topo}. In the SAT layer, we allow satellites in the same orbit to communicate with their adjacent neighbors (thus forming a ring too), but not those from different orbits. The reason is that satellites from different orbits have very high {\em relative velocity} and hence the impact of {\em Doppler shift} will become prominent and make communication unstable.

\subsection{Propagation of Local and Global Models}\label{sec:relay}

\begin{algorithm}
\caption{Propagation Algorithm of local and global models}\label{algorithm1}
\kwInit{epoch $\beta$ = 0, global model $w^{\beta}$, visible $\mathcal N_{s}$ = $\phi$}

\While{Stopping criterion not met }{
\ForEach(\Comment*[h]{\scriptsize propagate models via HAPs}){$h \in \mathcal H $}{
  \If{h is source HAP}{
   Transmit $w^{\beta}$ to its adjacent HAPs\\} 
  \ElseIf{h is sink HAP}{Stop relaying $w^{\beta}$\\ }
  \Else{Transmit $w^{\beta}$ to its next-hop HAP\\}

  Transmit $w^{\beta}$ to all its visible satellites $\mathcal N_{h}$\\
  Update $\mathcal N_{s} \leftarrow  \mathcal N_{s}$ $\cup$  $\mathcal N_{h}$\\
}  
\ForEach(\Comment*[h]{\scriptsize Propagate models via SATs}){$n \in  \mathcal N$}{   
\eIf{$n \in \mathcal N_s$}{
\Comment*[h]{\scriptsize Model propagation via visible SATs}\\
Transmit $w^{\beta}$ to its two neighboring satellites\\ 
Train $w^{\beta}$ to obtain $w_{n}^{\beta}$ \\
\eIf{$n$ still visible to h}{ 
Transmit $w_{n}^{\beta}$ and its metadata to $h$\\}
{Transmit $w_{n}^{\beta}$ and metadata to next-hop satellite}
}
{\Comment*[h]{\scriptsize invisible SAT}\\
Wait until $n$ receives $w^\beta$ from a neighbor\\
Train $w^{\beta}$ to obtain $w_{n}^{\beta}$\\
Transmit $w^{\beta}$, $w_{n}^{\beta}$, and metadata to next-hop satellite
}


}
{$ \beta \leftarrow  \beta+1$}
}
\end{algorithm}

We propose a model propagation algorithm that relays local and global models within the SAT layer and the HAP layer to speed up the FL training process under the severe constraint of highly intermittent satellite connectivity. 

\subsubsection{\bf Relaying global model in the HAP layer}\label{sec:rel_glb_hap}

When there are multiple HAPs, we pre-designate one HAP to be a {\em source} and another one to be a {\em sink} (typically the farthest from the source); they will also swap roles at appropriate times (see \sref{sec:rel_loc_hap}). The source HAP generates a global model $w^{\beta}$ (where $\beta=0$ means the initial global model) and transmits this model to its two adjacent HAPs via inter-HAP link (IHL). Each of these two HAPs will then pass $w^{\beta}$ to its next-hop neighbor (singular, since there is no need to send it back to the source). This relay continues on the ``ring'' until the model $w^{\beta}$ reaches the sink HAP, as illustrated in Fig.~\ref{Picture2}a. Along the way, each HAP will also broadcast the model $w^{\beta}$ to all of its visible satellites via the ``star'' topology\footnote{When there is only a single HAP, no model relay happens (just like the conventional GS case) and only the model broadcast will take place.}.

\subsubsection{\bf Relaying global and local models in the SAT Layer}

\begin{figure}[ht]
     \centering
     \includegraphics[width=1.0\linewidth]{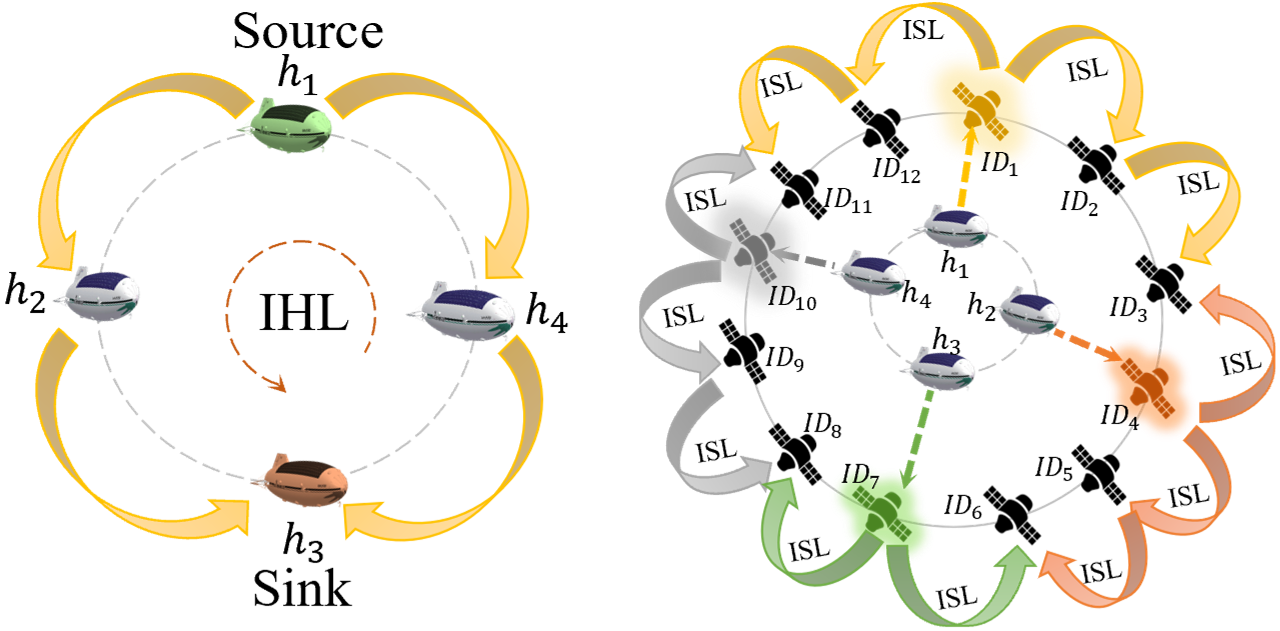}
     \begin{minipage}[t]{.5\linewidth}
        \subcaption{Model relay in the HAP layer.}
      \end{minipage}%
      \begin{minipage}[t]{.5\linewidth}
           \subcaption{Model relay in the SAT layer.}
      \end{minipage}
    \caption{Illustration of the proposed model propagation. The curved arrows represent model propagation directions. In (b), the four visible satellites $ID_{1-4-7-10}$ initiate the model relay, and the four satellites $ID_{3-6-8-11}$ cease the model relay.} 
    \label{Picture2}
\end{figure}

Once a visible satellite $n$ receives $w^{\beta}$ from a HAP, it performs two tasks simultaneously. One is to train $w^{\beta}$ using $n$'s local dataset $D_n$ to obtain an updated local model $w_{n}^{\beta}$, and the other is to send the global model $w^{\beta}$ to its neighboring satellites using ISL (without waiting for the model training task to complete). The second task is important because not all the satellites are visible to a HAP, and hence this model relay will kick start all the model training processes (each on a satellite) with {\em minimal delay}, using the latest version of the global model $w^{\beta}$. If a satellite receives the {\em same global model} from its two adjacent neighbors, the model relaying will cease at that satellite, as shown in Fig.\ref{Picture2}b.

Upon completion of the local training process, each satellite $n$ will send its trained local model $w_n^{\beta}$ to its currently visible HAP (random selection if multiple), together with some metadata (described in \sref{satellite_group}). If it does not have a visible HAP at the moment, it will send $w_n^{\beta}$ and metadata to its two adjacent satellites, who will either transmit the model to a HAP if they are visible to one; otherwise, it continues to relay $w_n^{\beta}$ to their respective neighbors too (but each to a single neighbor only because they would know the direction). This will substantially reduce the waiting time for each satellite to directly enter the visible zone of a HAP. 
\subsubsection{\bf Relaying local models in the HAP layer}\label{sec:rel_loc_hap}

Following the above process, each HAP will receive a set of local models $\{w_n^{\beta}\}$ with associated metadata from its visible satellites. Once this set reaches a certain point (determined in \sref{sec:group-agg}), the HAP will propagate these models to the sink HAP for model aggregation. This follows the same route shown in Fig. \ref{Picture2}a, from source to sink, but instead of relaying a global model, now each HAP is relaying a set of local models $\{w_n^{\beta}\}$ and metadata. Note that the set $\{w_n^{\beta}\}$ includes not only visible satellites' models but also some invisible ones', because of the local model relay in the SAT layer. Finally, when the sink HAP receives all the local models from other HAPs, it aggregates the local models 
into an updated global model $w^{\beta+1}$.

Next, the sink HAP will switch its role to a source, and propagates the updated global model $w^{\beta+1}$ to its neighbors until reaching the source HAP which has switched its role to a sink, following the reverse path of propagating $w^{\beta}$ as described in \sref{sec:rel_glb_hap}. Along the way, each HAP also transmits $w^{\beta+1}$ to its visible satellites. 

Algorithm \ref{algorithm1} summarizes the entire propagation process. Note that model training and relaying take place concurrently. 


\subsection{Convergence Operations} \label{sec:group-agg}

 
Without loss of generality, we assume satellite data are non-IID since they are collected from different orbits. As described in \sref{sec:relay}, the sink HAP will eventually collect all the local models and satellites' metadata from other HAPs, 
obtaining $\mathcal{U}=\bigcup_{h\in \mathcal H} u_{h}$, where $u_{h}$ contains all the local models and satellites' metadata collected by HAP $h$. However, simply aggregating these local models into a global model will result in poor performance because it fails to address three issues: (1) {\em model staleness} resulting from satellite models that were trained using an outdated global model,
(2) the {\em non-IID nature} of data collected from different orbits, and (3) the {\em data size variation} among visible satellites from different orbits. These will render the global model {\em biased} towards orbits that have frequently visible satellites and have larger data sizes.

Furthermore, unlike traditional asynchronous FL approaches, AsyncFLEO has another responsibility for each HAP to decide when to stop collecting the set of models $u_{h}$ and how to ``clean'' it, i.e., decide which orbits to include in the current global epoch (and what discount factor to apply), and which orbits to discard.

Therefore, AsyncFLEO performs two new functions: (1) grouping the satellites based on the diversity of their local models, and (2) aggregating the received local models in such a way that stale models do not adversely affect the FL model convergence.


\subsubsection{\bf Satellite Grouping}\label{satellite_group}

Once the sink HAP has collected all the local models, it organizes them as follows:
\begin{equation} \label{Eqn17}
    \mathcal{U}= \{ \mathcal S_{o_{1}}, \mathcal S_{o_{2}},\dots,\mathcal S_{\textit{O}}\},
\end{equation} 
where $\mathcal S_{o}$ is the set of all local models collected from orbit $o$ via HAPs $h_1,...,h_p$, which can be expressed as:
\begin{equation}\label{eqn_15}
    S_{o}=\{\underbrace{\{w_{n_1}^\beta,w_{n_2}^\beta,..,w_N^\beta\}_{h_1}}_{u_{h_1}},..,\underbrace{\{w_{n_1}^\beta,w_{n_2}^\beta,..,w_N^\beta\}_{H}}_{u_{H}}\}_{o}
\end{equation}
In addition, $\mathcal U$ also contains the {\em metadata} of each satellite $n$ in the collected set: a tuple $\langle ID, size, loc, ts, epoch \rangle_n$, where $ID$ is satellite $n$'s ID, $size$ is satellite $n$'s training data size, $loc$ is the satellite's current location (in an angular coordinate system) which is used to calculate its next visit time to PS, $ts$ is the time stamp when satellite $n$ transmits its local model to the PS, and $epoch$ is the last global epoch when satellite $n$ was included in updating the global model $w^{\beta}$ (i.e., if the latest $\beta=epoch$, then this local model is considered fresh).

Note that, each $\mathcal S_{o}$ and ultimately $\mathcal{U}$ could contain duplicate satellites' models and metadata, due to the possibility of some satellites being visible to more than one HAP at the same time. Thus, AsyncFLEO will filter out these duplicate models and obtain a cleaned set $\mathcal{U}$ which is composed of unique local models and their metadata
\big(i.e., $\{u_{hi}\}_{o} \cap \{u_{hj}\}_{o} = \phi$\big). Given the metadata, the sink HAP is able to determine the total data size of all the satellites in orbit $o$ as
\begin{equation}
    D_{\mathcal S_o}=\sum_{h=1}^{H} \sum_{n\in u_h} D_{n}
\end{equation}
where $D_{n}$ is the metadata $size$ of satellite $n$.


To deal with the straggler satellite problem and model staleness, AsyncFLEO groups satellites based on the similarity among their data distributions, which helps it to determine which satellites' models must be included in generating the global model and which could be discarded.

\textbf{Definition.} For a grouping $\mathcal G=\{\mathcal G_{1}, \mathcal G_{2}, \dots, \mathcal G_n\}$ over the set of all the satellites $\mathcal N$, a satellite $n\in \mathcal N$ is said to be {\em grouped} if $n\in\mathcal G_{i}$ for some group $\mathcal G_{i} \in \mathcal G$, and {\em ungrouped} if $n\notin \mathcal G_i$ for all groups $\mathcal G_{i} \in \mathcal G$. 
\begin{figure}[ht]
     \centering
     \includegraphics[width=1.0\linewidth]{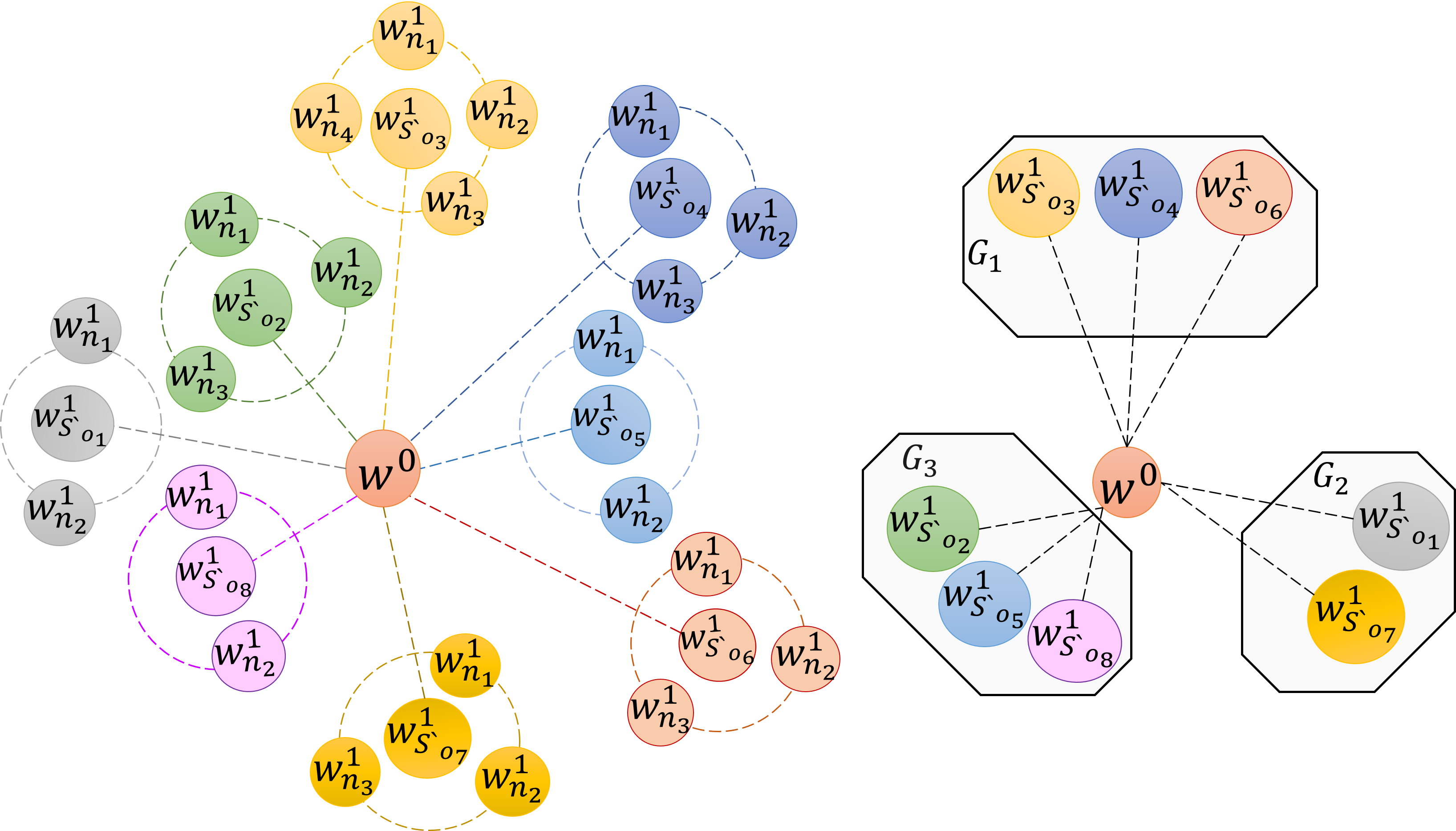}
     \begin{minipage}[t]{.5\linewidth}
       \centering
        \subcaption{Models before grouping.}
      \end{minipage}%
      \begin{minipage}[t]{.5\linewidth}
          \centering
           \subcaption{Models after grouping.}
      \end{minipage}
    \caption{Satellite (model) grouping of $O=8$ orbits into 3 groups $\mathcal G$. Dotted lines indicate the Euclidean distance between an orbit-wise aggregated model and the initial global model $w^0$.}
    \label{Picture4}
\end{figure}

In general, data collected by satellites in the same orbit tend to be similar to each other, while data collected from different orbits are more likely non-IID. This is due to the fact that satellites in the same orbit travel with the same orbital velocity $v_o$ (about 25,000 km/h) which is much faster than the rotation speed of the Earth (about 1,600 km/h); whereas, different orbits have different altitudes and inclination angles, thus resulting in different travel speeds and geographic areas covered by different orbits.

Since PS has no access to satellites' data as dictated by FL, AsyncFLEO groups satellites based on satellites' local models $w_n^{\beta}$ and the initial global model $w^{0}$. Specifically, during the first global epoch, local models generated by satellites from different orbits ($w_{n}^{1}$, $n=1,2,\dots, N$) are based solely on their local training dataset and are not affected (biased) by any other satellites' local models. Therefore, the weight divergence between each satellite's local model and the initial global model $w^{0}$ tends to be the greatest {\em in the first epoch}, thereby allowing for a most effective differentiation among satellites' models. This is also the reason why we choose $w^{0}$ instead of the latest global model $w^{\beta}$.


Based on this concept, the sink HAP first generates a {\em partial global model} $\mathcal S'_{o}$ for each orbit $o$ by performing a weighted average (according to data size) of the received models collected from orbit $o$ (see Fig.~\ref{Picture4}a: each partial global model is at the center of a dashed-line circle, while those on the circle are the ``member'' models participated in the weighted average). Next, it calculates the Euclidean distance between each $\mathcal S'_{o}$ and the initial global model $w^{0}$, $\|d_{ w_{\mathcal S'_o}^{1}} - d_{w^{0}}\|_{2}$. Those orbits with similar Euclidean distances to $w^{0}$ 
will be grouped together into a group $\mathcal G_{i}$ (see Fig.~\ref{Picture4}b). The sink HAP then stores this grouping scheme $\mathcal G=\{\mathcal G_{1}, \mathcal G_{2}, \dots, \mathcal G_n\}$ for use in subsequent epochs. 

In the next global epoch, when the sink HAP receives an updated version of satellites' models for any orbit, it checks whether that orbit is already in one of the stored groups. If so, this orbit of models will be directly assigned to the associated group. Otherwise, a partial global model will be computed on these models like the above, and its Euclidean distance to $w^{0}$ will again be computed, based on which this orbit will be assigned to the group that has the {\em minimum difference} from this orbit in terms of the average distance of its existing group members. The grouping procedure continues this way during each global epoch until all orbits have been grouped (hence typically taking only a few epochs). An illustration is given in Fig. \ref{Picture4}.

\begin{algorithm}[t]
\caption{\small{Model aggregation operation of AsyncFLEO } }\label{algorithm2}
   \kwInit{epoch $\beta=0$, global model $w^{\beta}$, $\mathcal G=\phi$}

\While{Termination criterion is not met}{
 Wait for receiving  $\mathcal S_o \hspace{0.1cm} \forall \hspace{0.1cm} o\in O$

{Build $\mathcal S_o  \leftarrow
 \bigcup_{u_{h_1}}^{u_H} \{w_{n_1}^\beta,\dots,w_N^\beta\}$}
 
 Update $\mathcal U$ using (\ref{Eqn17}) \\
 Filter $\mathcal U$ by removing redundant models \\
\ForEach (\Comment*[h]{\scriptsize grouping satellites}){$n \in \mathcal N$} { 

\If{$n$ \text{is not grouped}}{ 
{Compute $d_n= \|d_{w_n^0}-d_{w^0}\|_{2}$\\}
\ForEach (\Comment*[h]{\scriptsize compare similarity with grouped satellites}){$n' \in \mathcal G_i$ and all $i$} { 
{Assign $n$ to group $\mathcal G_i$ according to the similarity between $d_n$ and $d_{n'}$}}
}Update the grouping scheme $\mathcal G$}
%
\ForEach(\Comment*[h]{\scriptsize model aggregation}){$\mathcal G_i\in \mathcal G$}{
\If{none of the models in $\mathcal G_i$ is fresh}{
{Compute $\gamma$ for all $w_n^\beta \in \mathcal G_i$ using (\ref{eq:gamma})}}
\Else{\Comment*[h]{\scriptsize some or all are fresh}\\
Select fresh models $\{w_n^\beta\}$}
Generate $w^{\beta+1}$ using selected models via (\ref{AsyncFLEO}) }

 {$ \beta \leftarrow  \beta+1$}}
 
\end{algorithm}



\subsubsection{\bf Model Aggregation}

During each global epoch, AsyncFLEO selects which satellites' models should be included in generating the global model (i.e., model aggregation) and which should be excluded. The selection approach takes into account three main factors: (1) the staleness of each model (determined by its epoch as indicated by metadata $epoch$, relative to its group, (2) the number of satellites of each group, and (3) the total $size$ of all the satellites' data in each group. Based on these factors, and in principle, AsyncFLEO selects satellite models that are fresh and were trained with considerable data sizes to be included in model aggregation (i.e., generating the global model), while discarding stale models for this epoch only. The rationale is that the group possesses enough fresh models to compensate for the discarded stale models to participate in the global model aggregation. If, however, a group contains only stale models (no fresh models), AsyncFLEO will utilize these stale models but with a {\em staleness discounting factor} $\gamma$ for the group $\mathcal G_{i}$, defined as follows:
\begin{equation}\label{eq:gamma}
    \gamma= \sum_{ \mathcal G_i \in \mathcal G} \sum_{n\in \mathcal G_{i}}
    \bigg(\frac{D_{n}}{D}\bigg)  \bigg(\frac{k_{n}}{\beta}\bigg)
\end{equation}
where $n$ indexes the model using the pertaining satellite ID, ${D_{n}}/{D}$ is the ratio between the data size of this satellite $n$ and the total data size of all the satellites, and $k_{n}/\beta$ is the ratio between the last global epoch where satellite $n$ was included in generating the global model and the current global epoch $\beta$. 

Once the model selection has been completed, AsyncFLEO updates the global model as follows:
\begin{equation} \label{AsyncFLEO}
    w^{\beta+1}= (1-\gamma)  w^{\beta} + \sum_{g=\mathcal G_{1}}^ {\mathcal G_{n}}\sum_{n=1}^{N'_{g}} \gamma w_{n}^{\beta}
\end{equation}
where $N'_{g}$ is the total number of selected satellites from a group $g=G_i$. The rationale is that, when the current local models have grown stale, give the previous-round global model $w^{\beta}$ higher weight while local models lower weight, and vice versa. This way, the FL training and aggregation process repeats in each global epoch until reaching a termination criterion (e.g., a target accuracy or a maximum number of epochs) after multiple communication rounds (global epochs). As an optional step, the {\em final} FL model could be sent to a GS (e.g., by a HAP) if needed.

Algorithm \ref{algorithm2} summarizes the entire process of AsyncFLEO for grouping the satellites and generating the global model.





\section{Performance Evaluation}\label{sec:simu}


\subsection{Simulation setup}

\textbf{LEO Constellation.} We consider an LEO constellation consisting of 40 satellites equally distributed over five orbits. Each orbit is located at a height $h_o$ of 2000 km above the Earth's surface, with an inclination angle of 80$^\circ$. Two scenarios are considered for the PSs. One is a single GS or HAP located in Rolla, Missouri, USA (HAP floats above the city). The second scenario involves two HAPs, one floating above Rolla and the other above Portland, Oregon, USA. Each HAP hovers at an altitude of 20 km above the Earth's surface, with a minimum elevation angle $\vartheta_{min} = 10^\circ$ which is the same as the GS. A two-line element (TLE) set  \cite{web5} of each satellite is used by each PS to predict the satellite location on its trajectory. For each of the above two scenarios, the trajectories are determined over the course of three days. We select all communication link parameters discussed in Section~\ref{sec:model-links} to be consistent with the baselines that we compare with, as summarized in Table~\ref{Comm_links}.

\textbf{Baselines.} AsyncFLEO is compared to the most recent state-of-the-art methods, as reviewed in \sref{Sec:related_work}: FedISL~\cite{razmi} and FedHAP~\cite{happaper} which are synchronous approaches, and FedSat~\cite{razmi2022ground} and
FedSpace~\cite{so2022fedspace} which are asynchronous approaches.

\begin{table}[h]
\setlength{\tabcolsep}{1em}
\centering\arraybackslash
\renewcommand{\arraystretch}{1.5}
\caption{Simulation Parameters}
\label{Comm_links}
 \begin{tabular}{|p{4.3cm}|p{1.3cm}|} 
 \hline
   \textbf{Parameters} & \textbf{Values} \\
   \hline \hline
  Transmission power $P_{t}$& 40 dBm \\
  \hline
  Antenna gain of $G_{t},G_{r}$& 6.98 dBi\\
  \hline
 Carrier frequency $f$ & 2.4 GHz \\
  \hline 
 Noise temperature $T$& 354.81 K \\
  \hline 
 Transmission data rate $R$ & 16 Mb/sec   \\ 
  \hline  \hline
   Number of local training epochs $I$ & 100 \\
  \hline 
   Learning rate $\eta$ & 0.01 \\
  \hline 
   Mini-batch size $b_k$& 32\\
\hline 
\end{tabular}
\end{table}

\textbf{Dataset and ML models.} In line with most peer studies on FL-Satcom, we use the same two datasets in our evaluation. The first one is the MNIST dataset \cite{web2} which contains 70,000 grayscale images of handwritten digits of size 28$\times$28, and the other one is the CIFAR-10 dataset \cite{krizhevsky2009learning} which consists of 60,000 color images of 10 classes with a resolution of 32$\times$32 pixels. We consider both IID and non-IID data distributions among satellites. In the IID setting, training data samples are randomly shuffled and evenly distributed among all the satellites (each having all 10 classes of images). In the non-IID setting, satellites from two orbits have four classes of data, while satellites from the other three orbits have the remaining six classes. We consider two neural networks to train satellites: convolutional neural network (CNN) and fully connected multi-layer perceptron (MLP). The hyperparameters used for training are listed at bottom of Table~\ref{Comm_links}.
\begin{table}[t]

\setlength{\tabcolsep}{0.9em}
\centering\arraybackslash
\renewcommand{\arraystretch}{1.2}
\caption{Comparison with SOTA approaches}
\label{table1}

 \begin{tabular}{|p{1.5cm}|p{0.9cm} | p{1.3cm} | p{3.0cm}| } 

 \hline
 \rmfamily FL scheme & \centering Accuracy (\%) &  Convergence time (h:mm) & Remark \\

\hline  
 FedISL \cite{razmi} & 63.51 & 72& GS at arbitrary location \\
 \hline
 FedISL \cite{razmi} (ideal setup) & 81.74 & 3:30& PS is a GS at the NP or an MEO satellite above the Equator \\
 \hline
  FedSat \cite{razmi2022ground} (ideal setup) & 88.83 & 12 & GS at NP so that all satellites visit it at regular intervals \\
  \hline 
 FedSpace \cite{so2022fedspace} & 46.10 & 72 & GS needs satellites' local data \\
 \hline
FedHAP \cite{happaper} & 87.29 & 30 & HAP at arbitrary location \\
 \hline 
 \rowcolor{gray!20}
AsyncFLEO-GS & {\bf 80.62} & {\bf 6} &  GS at arbitrary location \\
  \hline 
  \rowcolor{gray!20}
AsyncFLEO-HAP & {\bf 81.36} & \textbf{5} & HAP at arbitrary location \\ 
  \hline 
  \rowcolor{orange!20}
\scriptsize AsyncFLEO-twoHAP & {\bf 82.94} & \textbf{3:20} & HAP at arbitrary location \\
 \hline
\end{tabular}
\end{table}

\begin{figure}[ht]
     \centering
     \includegraphics[width=1\linewidth]{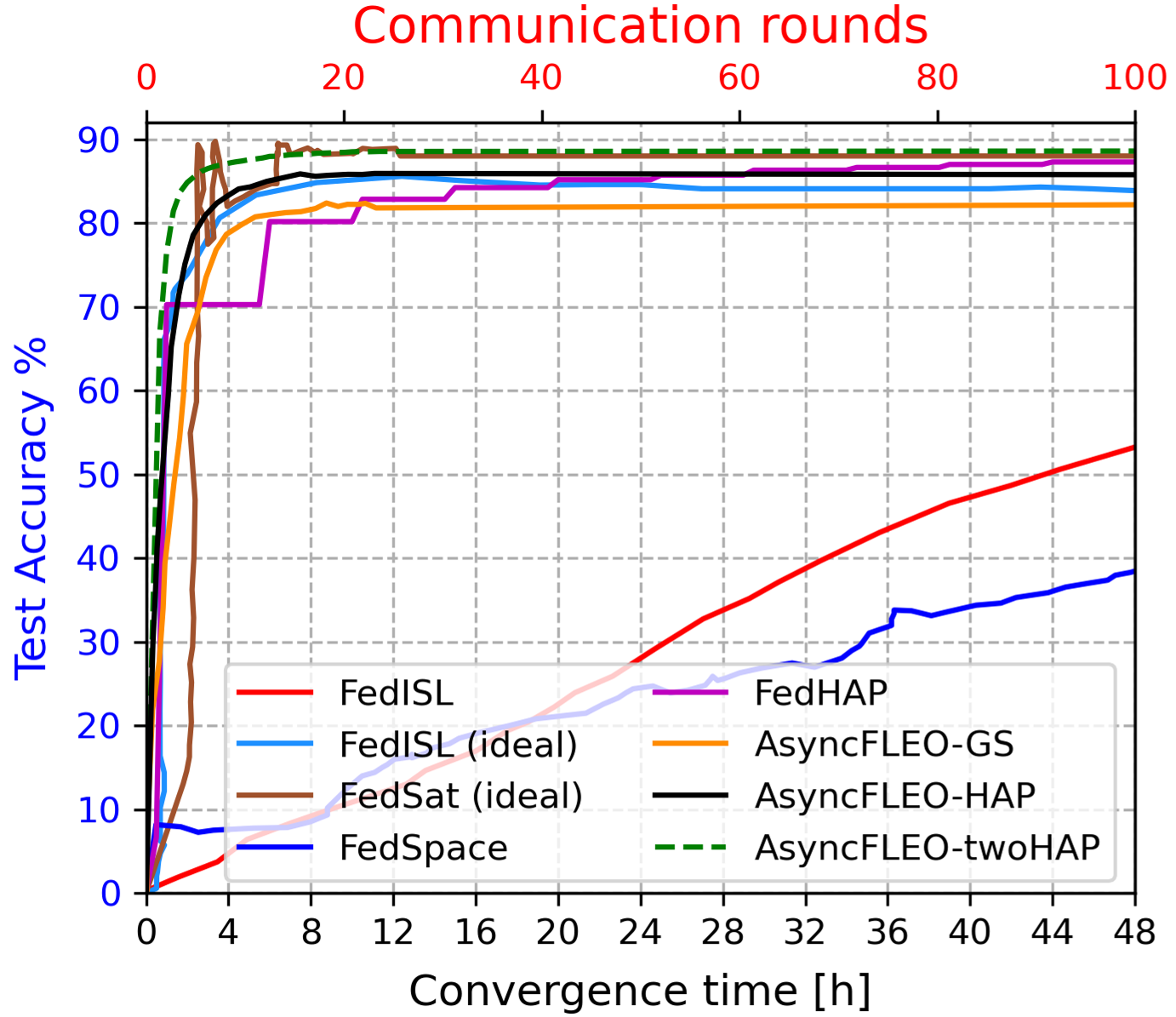}
    \caption{Accuracy vs. Convergence time: Comparison with state-of-the-art baselines using the MNIST dataset.}
    \label{compare}
\end{figure}
\begin{figure*}[t]
     \centering
     \begin{subfigure}[b]{0.325\textwidth}
         \centering
         \includegraphics[width=1\textwidth]{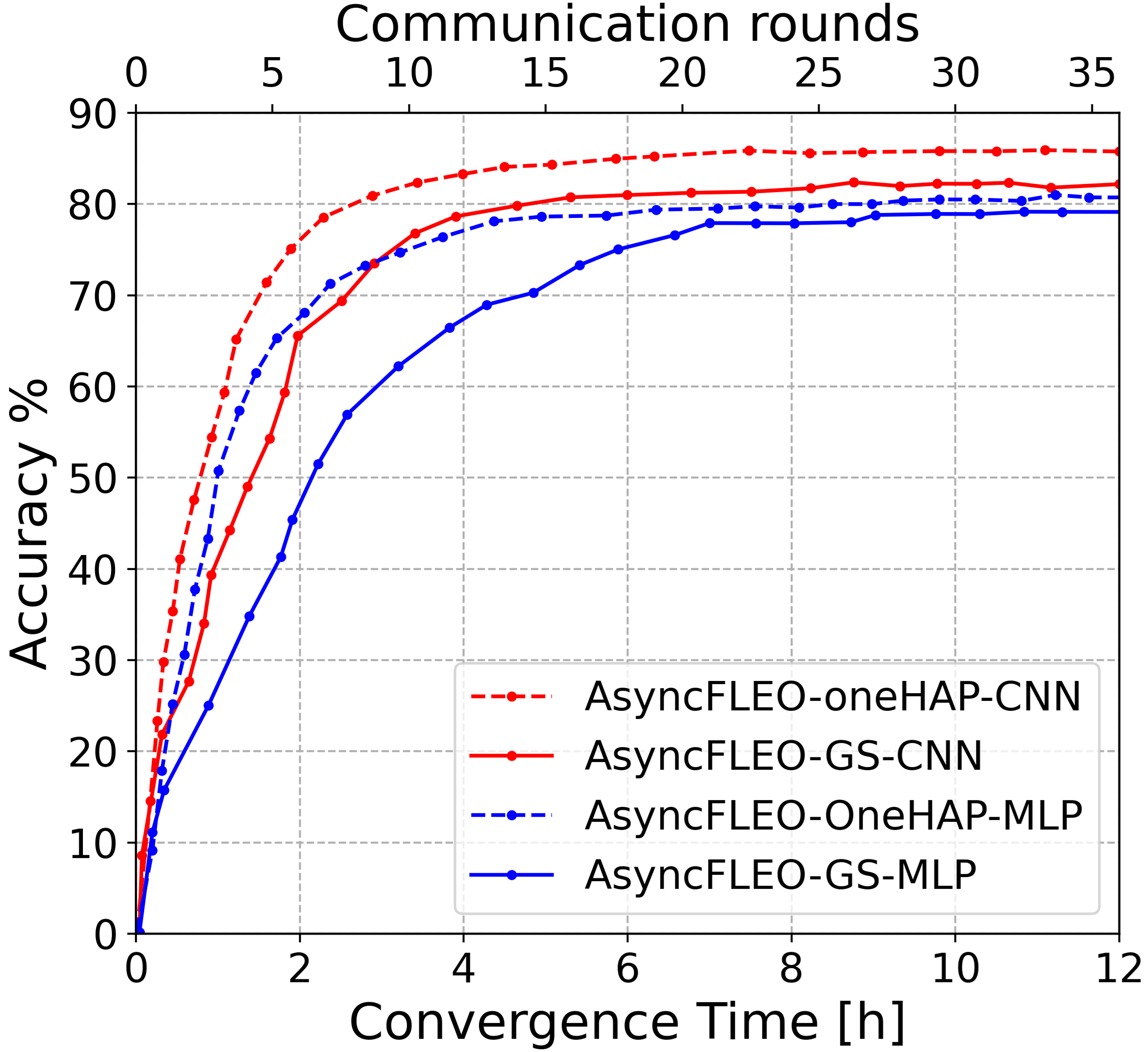}
         \caption{IID data.}
     \end{subfigure}
     \hfill
     \begin{subfigure}[b]{0.325\textwidth}
         \centering
         \includegraphics[width=1\textwidth]{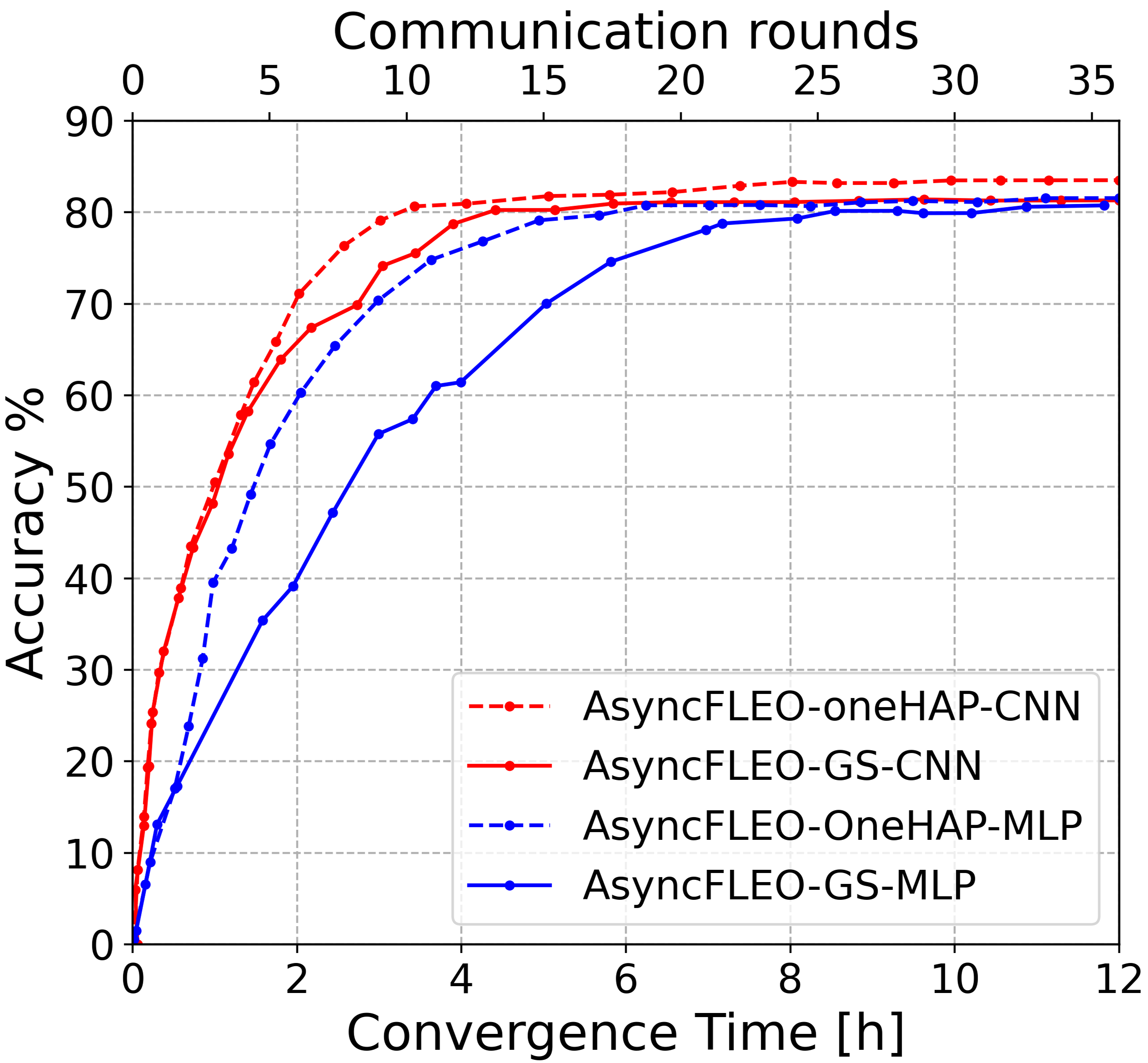}
         \caption{Non-IID data.}
     \end{subfigure}
     \hfill
     \begin{subfigure}[b]{0.325\textwidth}
         \centering
         \includegraphics[width=1\textwidth]{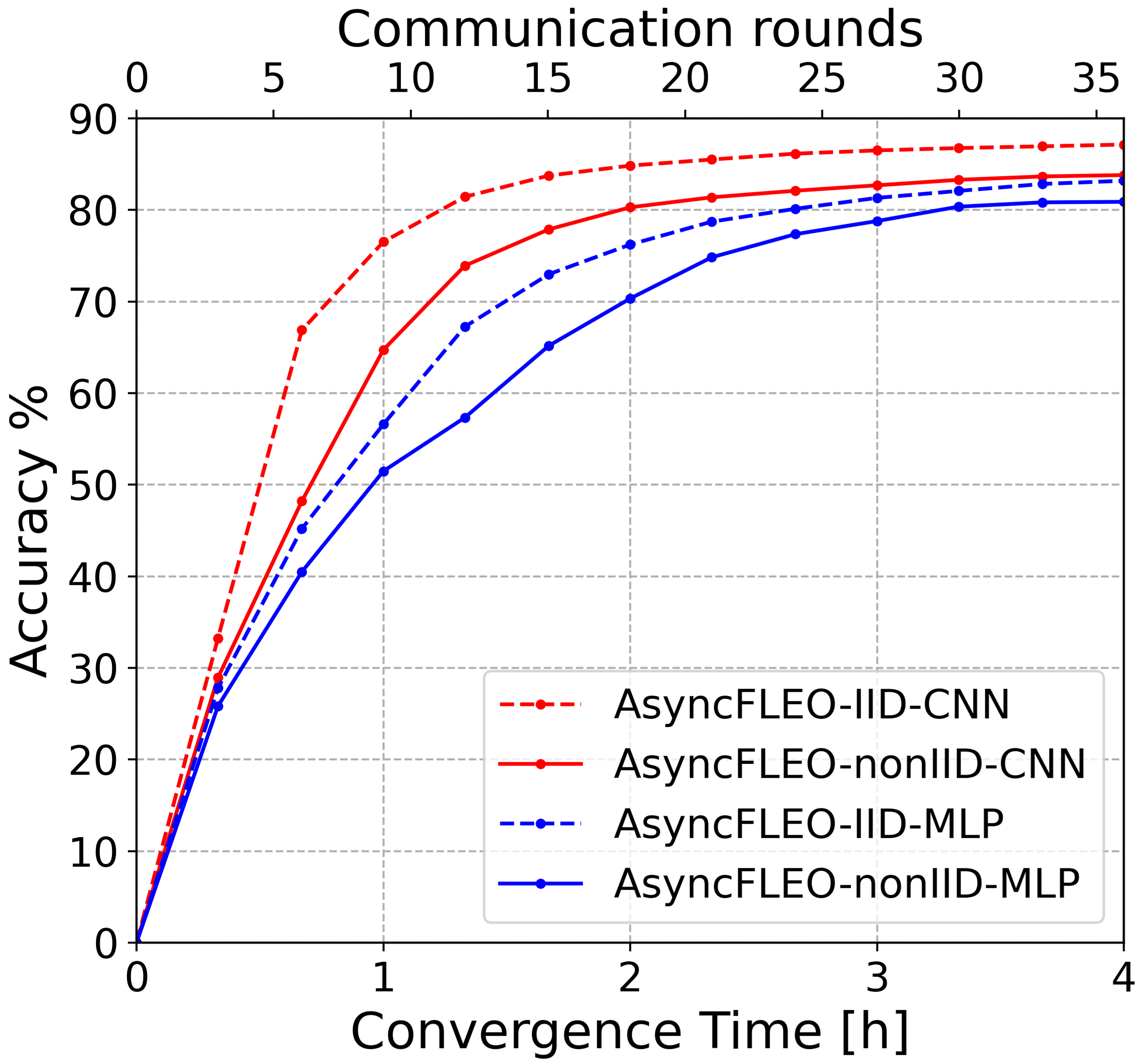}
         \caption{Two HAPs (Rolla and Portland).}
     \end{subfigure}
     \hfill
        \caption{AsyncFLEO evaluation on MNIST in various settings: IID/non-IID data, CNN vs. MLP, HAP vs. GS, one/two HAPs.}
        \label{fig:self-eval}
\end{figure*}

\begin{figure*}[t]
     \centering
     \begin{subfigure}[b]{0.325\textwidth}
         \centering
         \includegraphics[width=1\textwidth]{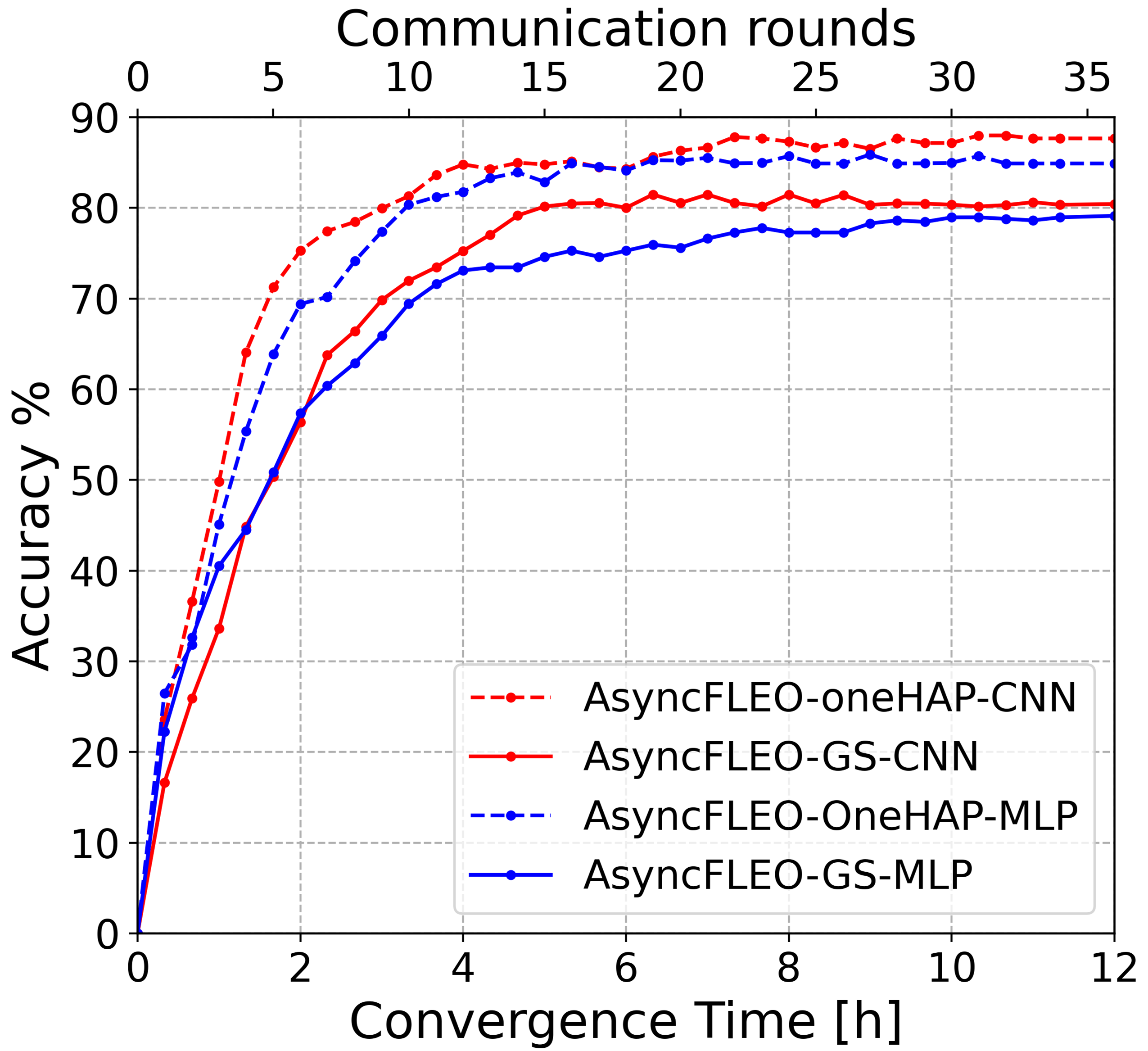}
         \caption{IID data.}
     \end{subfigure}
     \hfill
     \begin{subfigure}[b]{0.325\textwidth}
         \centering
         \includegraphics[width=1\textwidth]{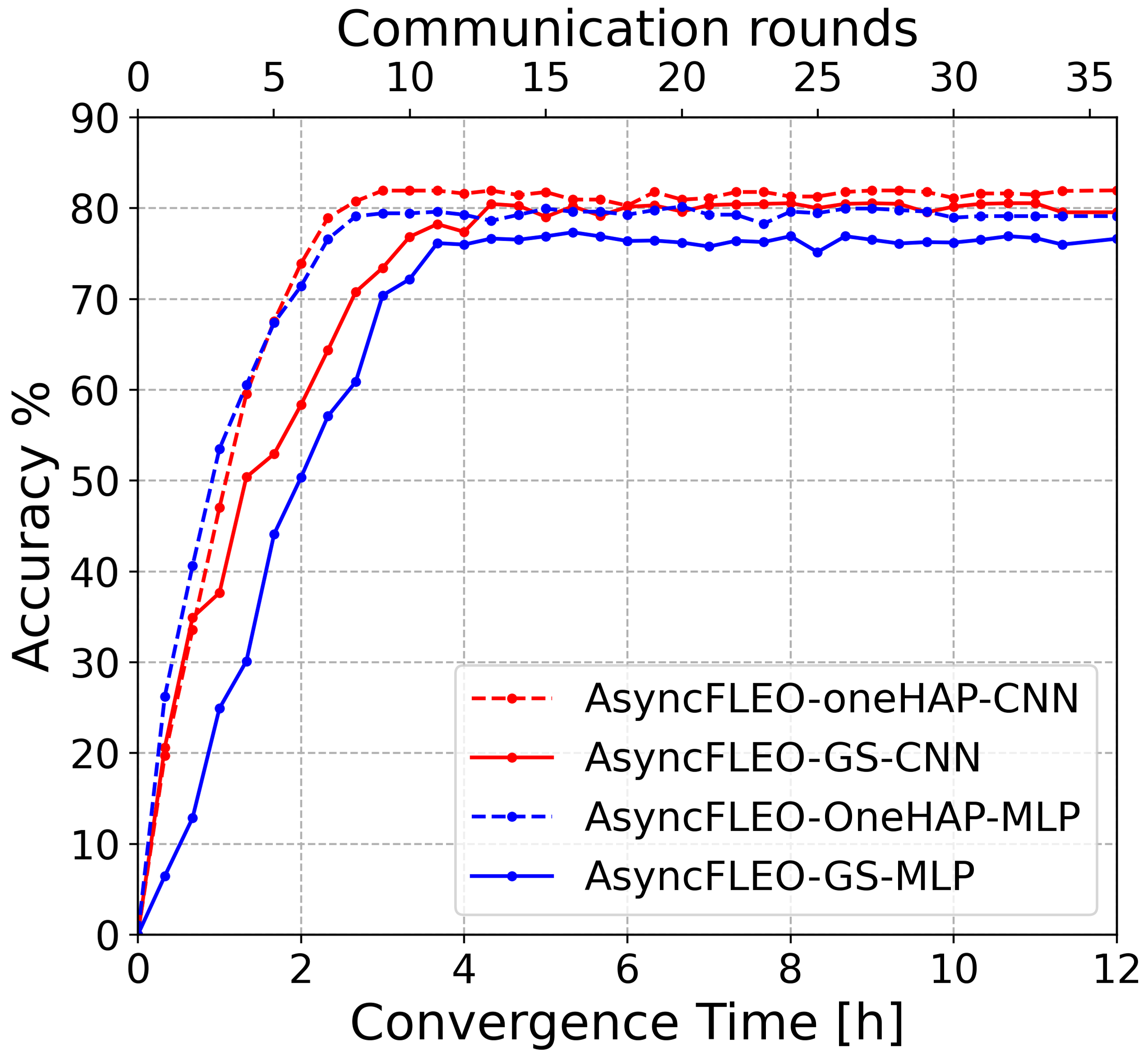}
         \caption{Non-IID data.}
     \end{subfigure}
     \hfill
     \begin{subfigure}[b]{0.325\textwidth}
         \centering
         \includegraphics[width=1\textwidth]{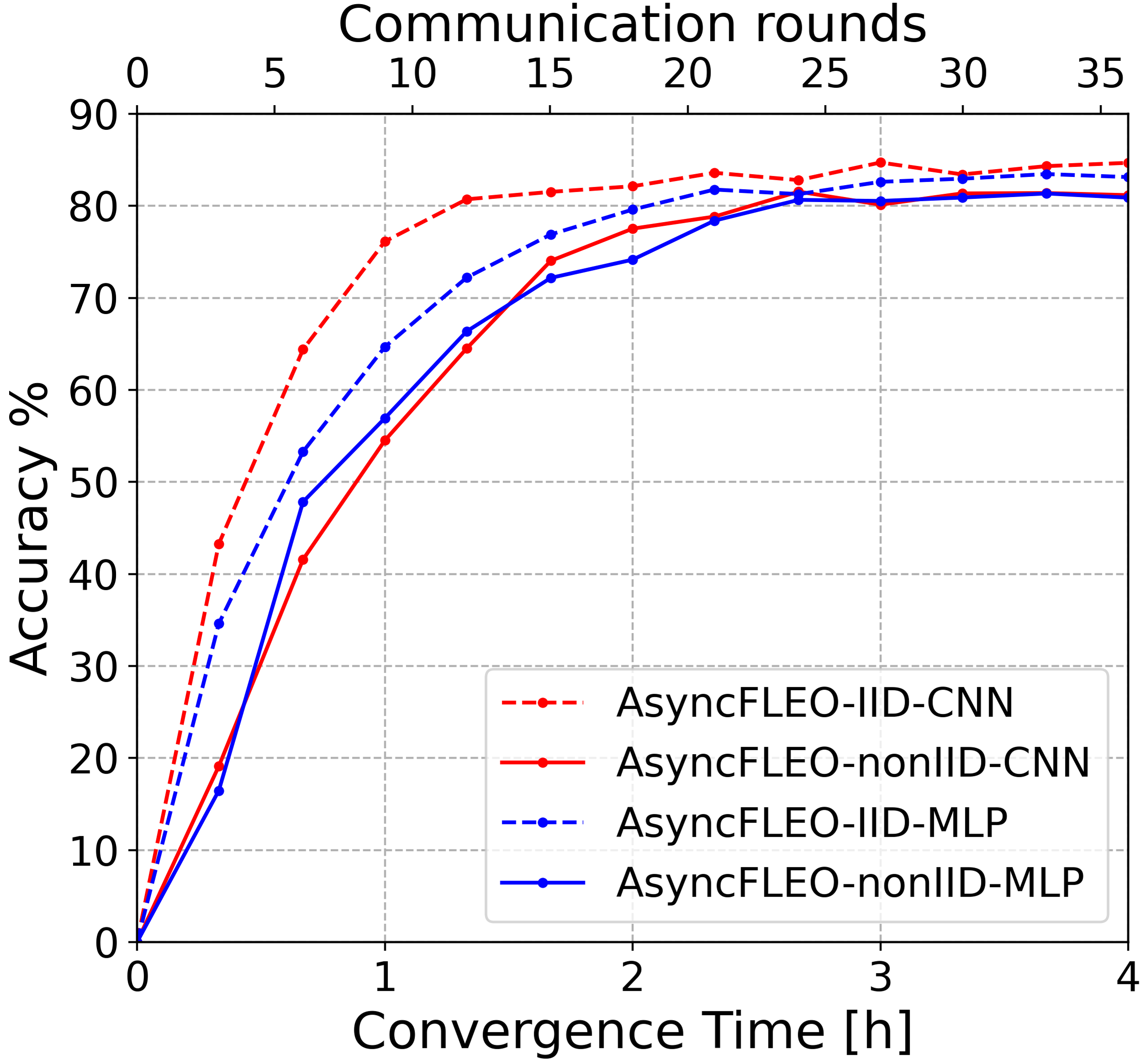}
         \caption{Two HAPs (Rolla and Portland).}
     \end{subfigure}
     \hfill
        \caption{AsyncFLEO evaluation on CIFAR-10 in various settings: IID/non-IID data, CNN vs. MLP, HAP vs. GS, one/two HAPs.}
        \label{fig:self-eval_2}
\end{figure*}

\subsection{Results}

\textbf{Comparison with State of the Art.}
Table \ref{table1} and Fig.~\ref{compare} summarize the comparison results on the MNIST dataset under the non-IID setting with CNN as a training network. Note that AsyncFLEO has two versions, AsyncFLEO-GS, and AsyncFLEO-HAP, since AsyncFLEO has the flexibility of using GS or HAP as its PS. For a fair comparison, AsyncFLEO-HAP uses a single HAP only (the multi-HAP case has better performance and is evaluated in the next subsection as shown in \fref{fig:self-eval}c and Fig.~\ref{fig:self-eval_2}c).

From Table \ref{table1} and Fig.~\ref{compare}, we observe that AsyncFLEO-twoHAP takes the shortest time 3:20 hours to converge and achieves a high accuracy of 82.94\%, and when it uses only a single HAP (for comparison with baselines), it achieves an accuracy of 81.36\% within only 5 hours. Although FedISL \cite{razmi} is faster than AsyncFLEO-HAP for convergent time 3:30 hours, it assumes an ideal setup (as discussed in Section\ref{Sec:related_work}), and even though given that condition, it only attains a lower accuracy of 81.7\%. 
When FedISL places the GS elsewhere, it takes as long as 72 hours and only achieves an accuracy of 63.5\%. While FedSat \cite{razmi2022ground} and FedHAP \cite{happaper} obtain slightly higher accuracy than AsyncFLEO-HAP, they take about 2.4 and 6 times longer convergence time, respectively, than ours to complete the learning process. In addition, FedSat \cite{razmi2022ground} assumes the similar ideal condition as FedISL \cite{razmi} which is very restrictive.

Among the two versions of AsyncFLEO, AsyncFLEO-HAP performs better than AsyncFLEO-GS as it is able to take advantage of the HAP's better satellite visibility due to its slightly elevated altitude. Nevertheless, AsyncFLEO-GS still performs fairly well, converging in just a few hours (6 hours) which is faster than most baselines and achieving a satisfactory accuracy (80.6\%). 


\textbf{Evaluating AsyncFLEO in more extensive settings.}
Here we evaluate AsyncFLEO more extensively under various settings: IID vs. non-IID data, CNN vs. MLP, and single- vs. multi-HAP, as shown in Fig.~\ref{fig:self-eval} and Fig.~\ref{fig:self-eval_2}, using MNIST and CIFAR-10 datasets, respectively. Fig. \ref{fig:self-eval}a gives the results obtained with the IID setting of the MNIST dataset. It shows that by just using a single HAP as the PS, AsyncFLEO can achieve an accuracy of 85.5\% within five hours, while AsyncFLEO using GS can also achieve an accuracy of 82.1\% after six hours. These are values when satellites employ CNN as their ML models. When using MLP, the accuracy slightly decreases by 4\% in both the GS and HAP cases, but they still converge within only a few hours. When the dataset changes from MNIST to CIFAR-10, AsyncFLEO again achieves good results. As shown in Fig.~\ref{fig:self-eval_2}a, AsyncFLEO attains an accuracy of 84.47\% in 4 hours and then 86.82\% after 8 hours using only a single HAP. This demonstrates AsyncFLEO's robustness to the change of datasets. 

Fig. \ref{fig:self-eval}b and Fig.~\ref{fig:self-eval_2}b demonstrate the robustness of AsyncFLEO to non-IID data settings. As can be observed from the two figures, AsyncFLEO converges in 12-15 asynchronous global epochs (four-six hours) with an accuracy of 79.67\%-81.36\% attained with a single HAP, and converges in 15-18 global epochs (five-six hours) with an accuracy of 78.43\%-80.15\% achieved with a GS. This difference comes again from the fact the HAP has better visibility of LEO satellites (about 1-5 satellites more at the same location) than the GS. When the training model changes from CNN to MLP, the accuracy only decreases negligibly while the convergence time increases by only 1-2 hours, which is still much more acceptable than those baseline methods. 

Finally, in Fig. \ref{fig:self-eval}c, we present the results for two HAPs, under both IID and non-IID data distributions. In the IID case, AsyncFLEO achieves an accuracy of 87.79\% in only 2:40 hours, while in the non-IID case, AsyncFLEO converges in 3:20 hours with an accuracy of 82.9\%. When CNN is substituted by MLP, the accuracy drops about 2-6\% for the IID case and the non-IID, respectively but is still above 80\% and the convergence time increased slightly to be around 6 hours, which is still rather desirable. When CIFAR-10 is used in place of the MNIST dataset as shown in Fig. \ref{fig:self-eval_2}c, the accuracy is slightly decreased and the convergence time is marginally increased to four hours, which is still significantly faster than the baselines. When we compare the results of one HAP with those of two HAPs (between (a) and (c) as well as between (b) and (c) in both Figs. \ref{fig:self-eval} and \ref{fig:self-eval_2}), it can be observed that the latter further speeds up the convergence process and improves the performance of FL.


\section{Conclusion} \label{section 4}
To usher FL into Satcom with maximal effectiveness and efficiency, we propose a novel asynchronous FL framework, AsyncFLEO, for LEO constellations. We address the challenges of highly sporadic connectivity and irregular visit patterns between satellites and PS, as well as model staleness caused by straggler satellites, which altogether lead to large convergence delays and poor model performance. AsyncFLEO groups satellites from different orbits based on the similarity of their data distribution inferred from model weights. Furthermore, AsyncFLEO introduces a ring-of-star communication topology and a model propagation algorithm. Our extensive simulations show that AsyncFLEO accelerates FL model convergence by up to 22 times and at the same time improves accuracy by up to 40\%, in comparison with several state-of-the-art approaches. The results also reveal that AsyncFLEO is robust to non-IID data, attaining an accuracy of 81.36\% in 5 hours which is quite on a par with its 85.57\% accuracy under the IID setting.

\bibliographystyle{IEEEtran}
\bibliography{references.bib}

\begin{thebibliography}{10}
\providecommand{\url}[1]{#1}
\csname url@samestyle\endcsname
\providecommand{\newblock}{\relax}
\providecommand{\bibinfo}[2]{#2}
\providecommand{\BIBentrySTDinterwordspacing}{\spaceskip=0pt\relax}
\providecommand{\BIBentryALTinterwordstretchfactor}{4}
\providecommand{\BIBentryALTinterwordspacing}{\spaceskip=\fontdimen2\font plus
\BIBentryALTinterwordstretchfactor\fontdimen3\font minus
  \fontdimen4\font\relax}
\providecommand{\BIBforeignlanguage}[2]{{%
\expandafter\ifx\csname l@#1\endcsname\relax
\typeout{** WARNING: IEEEtran.bst: No hyphenation pattern has been}%
\typeout{** loaded for the language `#1'. Using the pattern for}%
\typeout{** the default language instead.}%
\else
\language=\csname l@#1\endcsname
\fi
#2}}
\providecommand{\BIBdecl}{\relax}
\BIBdecl

\bibitem{wu2020resource}
H.~Wu, J.~Chen, C.~Zhou, W.~Shi, N.~Cheng, W.~Xu, W.~Zhuang, and X.~S. Shen,
  ``Resource management in space-air-ground integrated vehicular networks: Sdn
  control and ai algorithm design,'' \emph{IEEE Wireless Communications},
  vol.~27, no.~6, pp. 52--60, 2020.

\bibitem{perez2021airborne}
A.~Perez-Portero, J.~F. Munoz-Martin, H.~Park, and A.~Camps, ``Airborne gnss-r:
  A key enabling technology for environmental monitoring,'' \emph{IEEE Journal
  of Selected Topics in Applied Earth Observations and Remote Sensing},
  vol.~14, pp. 6652--6661, 2021.

\bibitem{fl2021}
P.~Kairouz, H.~B. McMahan, B.~Avent, A.~Bellet, M.~Bennis, A.~N. Bhagoji,
  K.~Bonawitz, Z.~Charles, G.~Cormode, R.~Cummings \emph{et~al.}, ``Advances
  and open problems in federated learning,'' \emph{Foundations and
  Trends\textsuperscript{\textregistered} in Machine Learning}, vol.~14, no.
  1--2, pp. 1--210, 2021.

\bibitem{so2022fedspace}
J.~So, K.~Hsieh, B.~Arzani, S.~Noghabi, S.~Avestimehr, and R.~Chandra,
  ``Fedspace: An efficient federated learning framework at satellites and
  ground stations,'' \emph{arXiv preprint arXiv:2202.01267}, 2022.

\bibitem{razmi}
\BIBentryALTinterwordspacing
N.~Razmi, B.~Matthiesen, A.~Dekorsy, and P.~Popovski, ``On-board federated
  learning for dense leo constellations,'' in \emph{IEEE International
  Conference on Communications (ICC)}, Seoul, Southkorea, May 2022. [Online].
  Available: \url{https://arxiv.org/abs/2111.12769}
\BIBentrySTDinterwordspacing

\bibitem{happaper}
\BIBentryALTinterwordspacing
M.~Elmahallawy and T.~Luo, ``{FedHAP}: Fast federated learning for {LEO}
  constellations using collaborative {HAPs},'' in \emph{2022 14th International
  Conference on Wireless Communications and Signal Processing}.\hskip 1em plus
  0.5em minus 0.4em\relax IEEE, Nov 2022. [Online]. Available:
  \url{https://arxiv.org/abs/2205.07216}
\BIBentrySTDinterwordspacing

\bibitem{hsieh2020uav}
F.~Hsieh, F.~Jardel, E.~Visotsky, F.~Vook, A.~Ghosh, and B.~Picha, ``Uav-based
  multi-cell haps communication: System design and performance evaluation,'' in
  \emph{GLOBECOM 2020-2020 IEEE Global Communications Conference}.\hskip 1em
  plus 0.5em minus 0.4em\relax IEEE, 2020, pp. 1--6.

\bibitem{arum2020review}
S.~C. Arum, D.~Grace, and P.~D. Mitchell, ``A review of wireless communication
  using high-altitude platforms for extended coverage and capacity,''
  \emph{Computer Communications}, vol. 157, pp. 232--256, 2020.

\bibitem{chen2022satellite}
H.~Chen, M.~Xiao, and Z.~Pang, ``Satellite-based computing networks with
  federated learning,'' \emph{IEEE Wireless Communications}, vol.~29, no.~1,
  pp. 78--84, 2022.

\bibitem{razmi2022ground}
N.~Razmi, B.~Matthiesen, A.~Dekorsy, and P.~Popovski, ``Ground-assisted
  federated learning in leo satellite constellations,'' \emph{IEEE Wireless
  Communications Letters}, 2022.

\bibitem{mcmahan2017communication}
B.~McMahan, E.~Moore, D.~Ramage, S.~Hampson, and B.~A. y~Arcas,
  ``Communication-efficient learning of deep networks from decentralized
  data,'' in \emph{Artificial intelligence and statistics}.\hskip 1em plus
  0.5em minus 0.4em\relax PMLR, 2017, pp. 1273--1282.

\bibitem{walker1984satellite}
J.~G. Walker, ``Satellite constellations,'' \emph{Journal of the British
  Interplanetary Society}, vol.~37, p. 559, 1984.

\bibitem{xie2020asynchronous}
C.~Xie, O.~Koyejo, and I.~Gupta, ``Asynchronous federated optimization,'' in
  \emph{12th OPT Workshop on Optimization for Machine Learning}, 2020.

\bibitem{web5}
``Nasa, definition of two-line element set coordinate system [online],''
  Available: \url{https://https://tinyurl.com/yc36cwju}, May 2022.

\bibitem{web2}
L.~Deng, ``The mnist database of handwritten digit images for machine learning
  research,'' \emph{IEEE Signal Processing Magazine}, vol.~29, no.~6, pp.
  141--142, 2012.

\bibitem{krizhevsky2009learning}
A.~Krizhevsky, V.~Nair, and G.~Hinton, ``Cifar-10 (canadian institute for
  advanced research),'' \emph{http://www. cs. toronto. edu/kriz/cifar. html},
  vol.~5, no.~4, p.~1.

\end{thebibliography}

\end{document}